\documentclass[letterpaper, 10 pt, conference]{ieeeconf}  

\IEEEoverridecommandlockouts                              

\usepackage{booktabs}
\usepackage{graphics} 
\usepackage{graphicx}
\usepackage{epsfig} 
\usepackage{mathptmx} 
\usepackage{multirow}
\usepackage{times} 
\usepackage{amsmath} 
\usepackage{amssymb}  
\usepackage{xcolor} 
\usepackage{tabularx}
\usepackage{cite}
\usepackage{ulem}
\usepackage{float}
\useunder{\uline}{\ul}{}
\usepackage{hyperref}

\title{\LARGE \bf
CORENet: \uline{C}ross-M\uline{o}dal 4D \uline{R}adar D\uline{e}noising \uline{Net}work with LiDAR Supervision for Autonomous Driving
}

\author{Fuyang Liu$^{1,2}$, Jilin Mei$^{1}$, Fangyuan Mao$^{1,2}$, Chen Min$^{1}$, Yan Xing$^{3}$, and Yu Hu$^{1,2}$
\thanks{*This work was supported by National Natural Science Foundation of China under Grant No.U23B2034, No.62203424, and No.62176250: and the Innovation Program of Institute of Computing Technology, Chinese Academy of Sciences under Grant No. 2024000112.}
\thanks{$^{1}$Institute of Computing Technology, 
                Chinese Academy of Sciences, China,
                \{liufuyang2023, meijilin, maofangyuan23s, mincheng, huyu\}@ict.ac.cn
        }%
\thanks{$^{2}$University of Chinese Academy of Sciences, China}%
\thanks{$^{2}$Beijing Institute of Control Engineering, China, xingyan@bice.org.cn}
}

\begin{document}

\maketitle
\thispagestyle{empty}
\pagestyle{empty}

\begin{abstract}
4D radar-based object detection has garnered great attention for its robustness in adverse weather conditions and capacity to deliver rich spatial information across diverse driving scenarios. Nevertheless, the sparse and noisy nature of 4D radar point clouds poses substantial challenges for effective perception. To address the limitation, we present CORENet, a novel cross-modal denoising framework that leverages LiDAR supervision to identify noise patterns and extract discriminative features from raw 4D radar data. Designed as a plug-and-play architecture, our solution enables seamless integration into voxel-based detection frameworks without modifying existing pipelines. Notably, the proposed method only utilizes LiDAR data for cross-modal supervision during training while maintaining full radar-only operation during inference. Extensive evaluation on the challenging Dual-Radar dataset, which is characterized by elevated noise level, demonstrates the effectiveness of our framework in enhancing detection robustness. Comprehensive experiments validate that CORENet achieves superior performance compared to existing mainstream approaches. The code is available at \url{https://github.com/charlesuv/corenet.git}.

\end{abstract}

\section{INTRODUCTION}
\begin{figure}[htbp]
  \centering
  \includegraphics[width=1\linewidth]{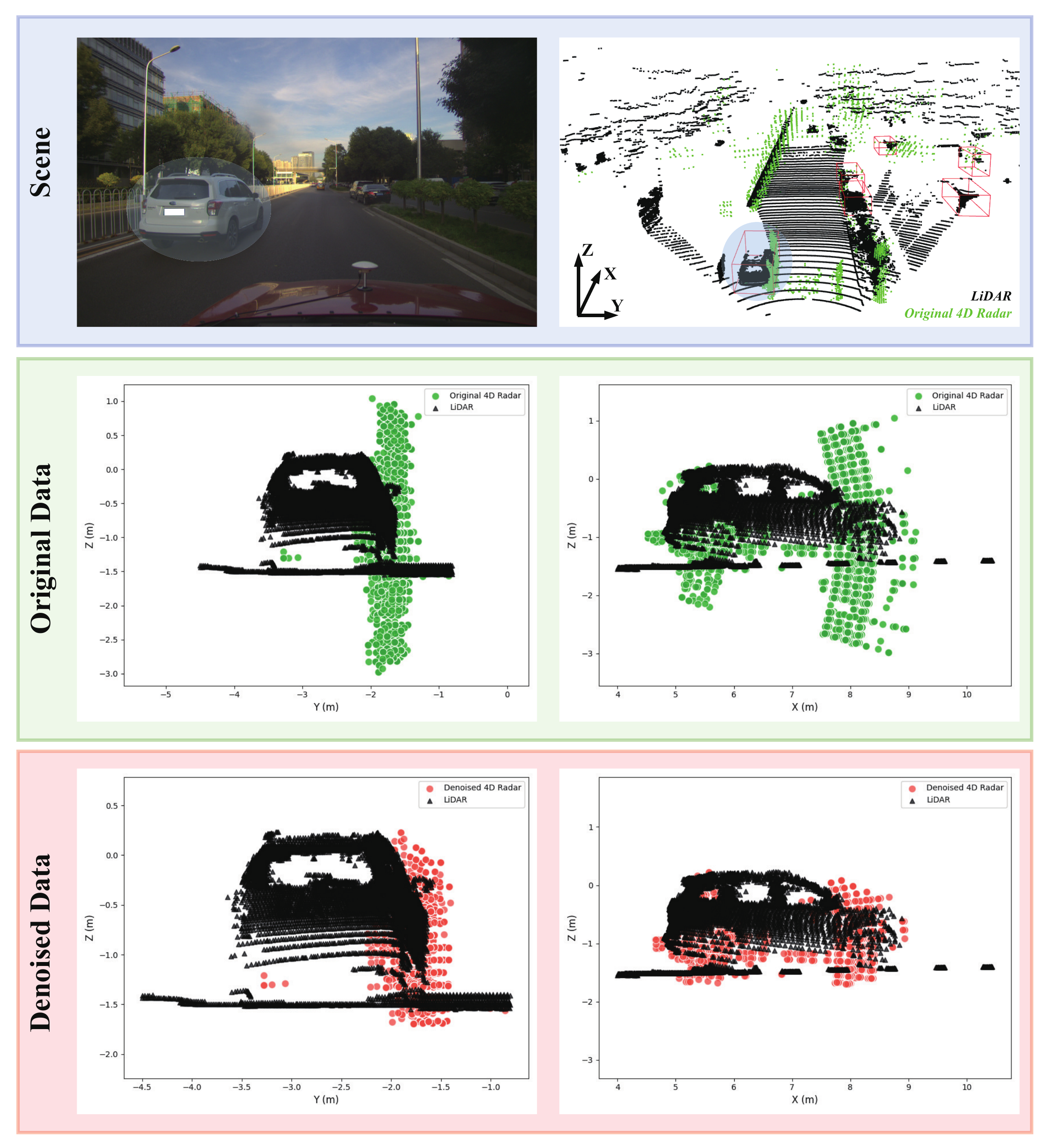}
  \caption{Illustration of 4D millimeter-wave radar with extensive noise and LiDAR point clouds. The first row is a scene overview where green and black point clouds represent raw 4D radar and LiDAR data, respectively. A nearby car is annotated with a blue circle. The last two rows zoom into the blue circle region. The second row shows the original radar data with green points. The third row is the denoised results using our method, indicating the effectiveness of noise suppression in the data processing.}
  \label{fig:vis_noise}
\end{figure}
Environmental perception serves as the fundamental module of autonomous driving systems, providing essential support for downstream modules, including path planning, decision-making, and vehicle control~\cite{cui2021deep}. While cameras and LiDAR currently serve as the primary sensor modalities in most autonomous driving platforms, their performance is not ideal in adverse weather conditions~\cite{yang2020radarnet}. Meanwhile, the radar has gained significant attention due to better environmental adaptability and reliability attributed to their use of relatively longer wavelengths~\cite{Abdu2021ApplicationOD}. In the field of autonomous driving, mainstream radar sensors are mainly divided into two categories: 3D radar and 4D radar~\cite{zhang2023dual}. Traditional 3D radar suffers from limited resolution and the absence of elevation measurement, which impede precise object characterization and restrict the deployment in autonomous driving scenarios. In contrast, 4D imaging radar provides enhanced spatial perception through four-dimensional data acquisition (range, azimuth, elevation, and Doppler velocity), delivering comprehensive environmental awareness essential for robust operation in complex driving conditions. 

However, point clouds collected by radar exhibit strong noise, which is especially prominent in 4D radar, as shown in Fig.~\ref{fig:vis_noise}. The strong noise increases the false detection rate. Accurately distinguishing valid and noise points in 4D radar point clouds poses a significant challenge for 4D radar perception strategies. In the early stages, plenty of radar denoising methods are proposed for 3D radar. Some traditional optimization-based approaches like Moving Least Squares and low-rank learning~\cite{amenta2004defining, fleishman2005robust, chen2019multi} leverage geometric projections and self-similarity but often struggle with complex, noisy, and highly irregular radar data. Worth mentioning, neural network-based methods~\cite{yu2022point,du2022dncnet} achieve impressive results in low-noise conditions but face challenges with denser and more variable noise levels, as well as larger-scale point clouds typical in radar data. In radar signal processing, denoising is commonly performed using radius-based~\cite{shamsfakhr2024multi} or statistics-based~\cite{xu2023mmlock} methods, which can partially eliminate outlier noise but fail to identify extensive noise distributions.

Upon analyzing the noise characteristics of 4D radar and reviewing the fundamental principles of radar signal imaging \cite{sun20214d,liu2024planar}, we identify that the primary noise source arises from energy leakage in sidelobes surrounding strong targets. This phenomenon is predominantly caused by the inherent limitations in the resolution of radar systems \cite{sun20214d}. It is known from~\cite{zhang2023dual} that the radar has a low resolution in the vertical direction, which results in the distribution of noise along the vertical tangent direction in Fig.~\ref{fig:vis_noise}. Therefore, improving the perceptual capability of 4D radar requires identifying the points associated with actual objects in the point cloud. Drawing inspiration from recent advancements in multi-modal fusion~\cite{yan20222dpass, zhang2023delivering, bang2024radardistill}, we try to improve the detection capability of 4D radar-based models by constraining the distribution of 4D radar data with LiDAR. Some works~\cite{yang2020radarnet,palmer2024lerojd} focus on modal fusion and attempt to mitigate radar’s limitations by integrating LiDAR information, but they remain largely unexplored for 4D radar, where noise and sparsity are even more pronounced, limiting their robustness and generalizability in real-world scenarios. We propose utilizing cross-modal learning, which leverages LiDAR point clouds during training to make the detection model implicitly capable of recognizing noise. 

In this paper, we introduce a cross-modal 4D radar supervision framework CORENet, which effectively distinguishes noise and valid point clouds by using LiDAR point clouds to align with 4D radar point clouds. Experimental results demonstrate that point clouds supervised across modalities can better fit target objects, thereby enhancing object detection capabilities. 

In summary, the contributions can be outlined as follows: \\
\begin{itemize}
\item[$\bullet$] A hierarchical multi-scale denoising module named HMSD-Net for 4D radar is proposed, which is simple and flexible, allowing seamless integration into a voxel-based object detection network to improve detection accuracy.
\item[$\bullet$] We employ a cross-modal supervision mechanism that utilizes LiDAR data to supervise the training of 4D radar data. Our approach requires LiDAR point clouds in the training stage only, ensuring simplicity and efficiency during the inference stage. 
\item[$\bullet$] Experimental results show that this supervised denoising framework enables the radar point cloud to better fit target objects and improves the mean accuracy of the original model by 7\%-18\% in 3D and BEV views. 
\end{itemize}

\section{RELATED WORKS}
\begin{figure*}[t]
  \centering
  \includegraphics[width=1\linewidth]{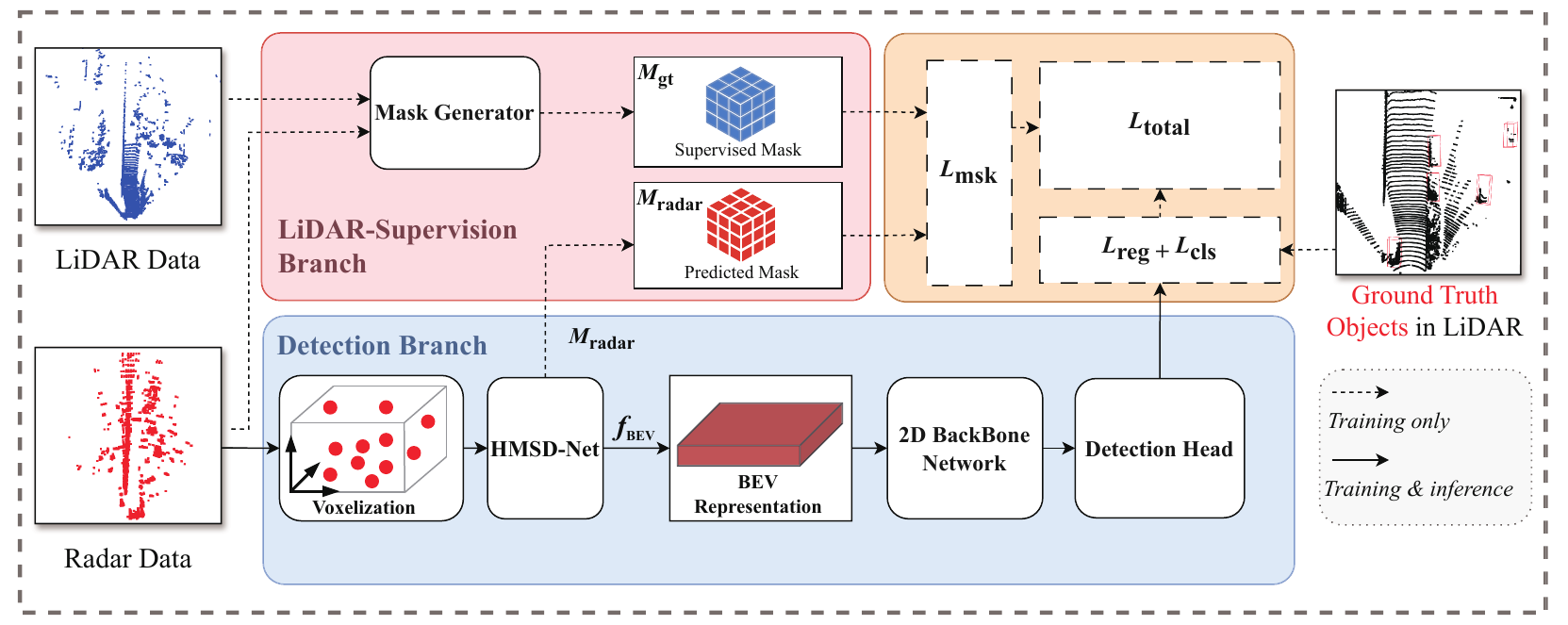}
  \caption{The overview of our proposed CORENet. LiDAR point clouds are processed through the Mask Generator to produce a Supervised Mask, which serves as supervision for radar point clouds. The radar point clouds are simultaneously passed through the voxelization operation and HMSD-Net, where the radar data undergoes feature coding to suppress the noise features and thus achieve the denoising effect. Then, the denoised radar data is passed through the 2D Backbone Network, followed by the Detection Head. The training process is optimized by computing the total loss with a voxel-level loss computed.}
  \label{figurelabel}
\end{figure*}

\subsection{Radar-based Object Detection}
Radar-based 3D object detection models adopt backbone networks initially designed for LiDAR detectors, tailored specifically to meet the unique requirements of radar data. RadarPointGNN~\cite{svenningsson2021radar} employs Graph Neural Networks (GNNs)~\cite{shi2020point} to effectively extract features from sparse radar point clouds through graph representations. KPConv Pillars~\cite{ulrich2022improved} introduces a hybrid architecture that combines grid-based and point-based approaches for radar-only 3D object detection. Recent studies have increasingly focused on detection techniques that fuse radar with LiDAR or cameras~\cite{kim2023craft, kim2023crn, nabati2021centerfusion, wang2023bi, yang2020radarnet, zhou2023bridging}, aiming to leverage radar data to compensate for the limitations of each sensor. RadarNet~\cite{yang2020radarnet} adopts a voxel-based early fusion and an attention-based late fusion approach to integrate radar and LiDAR data. RCM-Fusion~\cite{kim2024rcm} fuses radar and camera data at both the feature and instance levels, maximizing the potential of radar information. CRN~\cite{kim2023crn} leverages radar data to transform camera image features into a bird’s-eye view (BEV) and then combines these transformed camera features with radar BEV features using a multimodal deformable attention mechanism.

\subsection{Radar Denoising}
Due to the high noise and excessive sparsity of millimeter-wave radar point clouds, perception solutions based on such data are typically underutilized. As a fundamental challenge, point cloud denoising has garnered significant attention. Initially, more traditional approaches such as Moving Least Squares (MLS)~\cite{oztireli2009feature,alexa2003computing} were used to project point sets onto effective planes. However, these classical methods were originally designed for reconstructing noise-free surfaces and are not well-suited for describing complex polygonal objects. Moreover, these traditional approaches often involve complex computations and iterative processes, limiting their broader applicability. In radar signal processing, mainstream denoising methods predominantly employ Radius Outlier Removal (ROR)~\cite{shamsfakhr2024multi} and Statistical Outlier Removal (SOR)~\cite{xu2023mmlock}. These approaches are widely used due to their excellent efficacy in mitigating isolated outlier noise while exhibiting limited capability in addressing complex spatial noise distributions.

In recent years, with the rapid advancement of neural networks~\cite{wang2019dynamic, qi2017pointnet, qi2017pointnet++, liu2021deep, de2023iterativepfn}, deep learning techniques have been extensively applied to point cloud denoising, yielding remarkable results. Point-BERT~\cite{yu2022point} employs a masked learning strategy to restore incomplete point clouds, enhancing their noise resilience. However, this method primarily handles lower noise levels and is suited for smaller-scale point cloud data. DNCNet~\cite{du2022dncnet} uses a U-shaped denoising sub-network to eliminate signal noise, followed by a pre-trained classification sub-network to identify the denoised signal types, but its application is limited to one-dimensional signals with simpler forms and features. Furthermore, there are efforts to integrate LiDAR and radar data, such as RadarNet~\cite{yang2020radarnet}, which fuses LiDAR data with radar data from a bird’s-eye view (BEV) and utilizes velocity information to improve object detection accuracy. RadarDistill~\cite{bang2024radardistill} applies a distillation technique to constrain radar feature representations using LiDAR data. Nevertheless, existing methods have primarily been evaluated on 3D radar systems, while their effectiveness in 4D radar with elevated noise levels remains unexplored.
\section{METHOD} \label{sec:method}
The proposed method introduces a novel denoised framework for radar point clouds in 3D object detection task, using LiDAR point clouds to supervise the denoising process of 4D radar data during training only while eliminating the need for LiDAR data during inference, as shown in Fig.~\ref{figurelabel}.

\subsection{Hierarchical Multi-Scale Denoising Network} \label{sub:hmsdnet}
\begin{figure}[t]
  \centering
  \includegraphics[width=0.98\columnwidth]{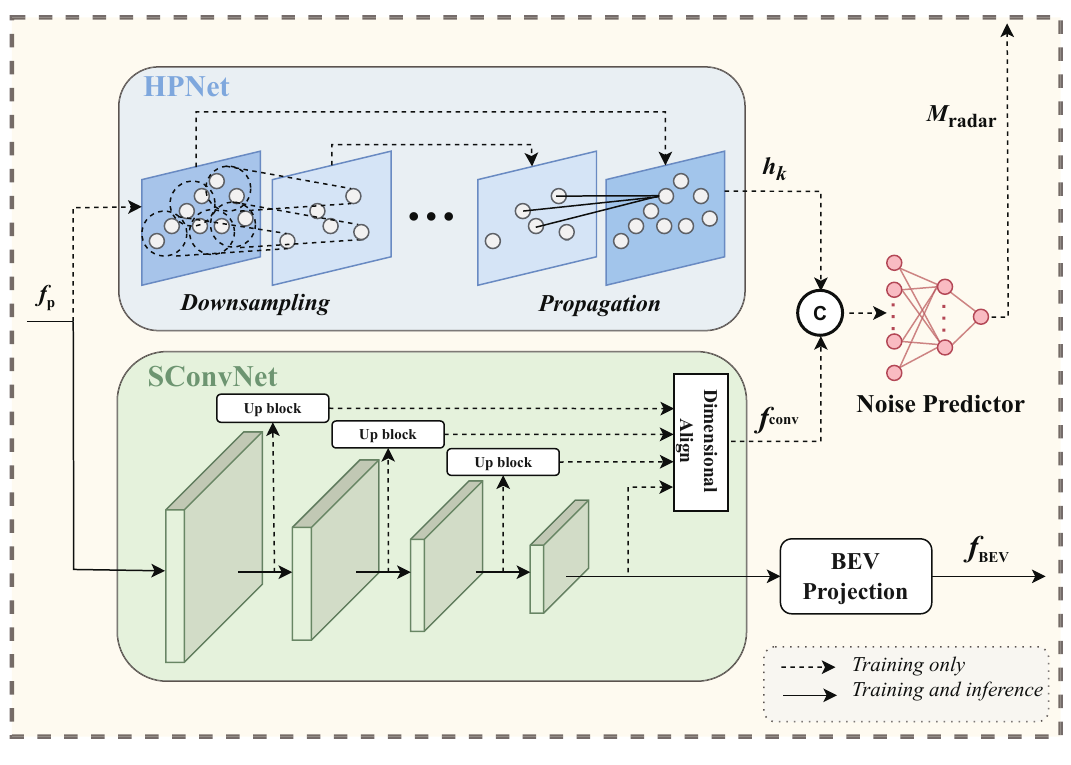}
  \caption{The structure of HMSD-Net (see Sec.~\ref{sub:hmsdnet} for details)}
  \label{figurelabel_sub}
\end{figure}

 As illustrated in Fig.~\ref{figurelabel_sub}, Hierarchical Multi-Scale Denoising Network (HMSD-Net) integrates multi-scale features from two key components: (1) Hierarchical PointNet (HPNet) for topological feature learning and (2) Sparse ConvNet (SConvNet) for spatial encoding. The multi-scale features are later output through a network named Noise Predictor to obtain noise predictions.

Based on the visualization in Fig.~\ref{fig:vis_noise} and principles of radar imaging~\cite{sun20214d}, the sidelobe signals distributed around targets contain valuable structural information, which can be effectively enhanced through multi-scale local feature aggregation. This signal reinforcement mechanism improves the detection capability in complex environments by preserving fine-grained spatial relationships while suppressing interference patterns. Voxel convolution can capture local spatial information but fails to preserve topological relationships between points~\cite{deng2021voxel}. 

We draw inspiration from works of graph convolution~\cite{qi2017pointnet,zhou2021adaptive} to design the HPNet. The HPNet applies $k$ encoding layers that progressively extract aggregated features from voxels.
Specifically, in downsampling layers, the farthest point sampling (FPS) is applied, followed by k-nearest neighbor clustering to aggregate voxels. In propagation layers, the clustered voxels undergo non-linear transformations through cascaded multi-layer perceptrons (MLPs) with shared weights, where the MLP layers propagate features from sparse layers to dense layers. These adjacent connections with skip concatenation capture complex interactions within the latent feature representations \{$h_1, h_2, ..., h_k$\}.

Concurrently, the SConvNet processes voxel features through a multi-stage architecture comprising submanifold sparse convolution layers~\cite{graham20183d}, batch normalization, and ReLU activations at each hierarchy. Specifically, each processing block applies submanifold sparse convolution to maintain geometric sparsity, followed by normalization and ReLU to stabilize feature distributions. The hierarchical features from SConvNet undergo dimensional alignment via upsampling operations, followed by concatenation to synthesize the composite feature representation $f_{\text{conv}}$.

Given the radar point voxels $f_{\text{p}} = \{p_1, p_2, ..., p_N\}$, where each voxel $p_i \in \mathbb{R}^{d}$ encapsulates raw radar information: 3D coordinates $[x, y, z]$, signal intensity $i_s$, and radial Doppler velocity $ v_D $, the hierarchical processing is formalized as:
\begin{equation}
\begin{gathered}
    h_k = \textbf{HPNet}(f_{\text{p}}),\  h_k \in \mathbb{R}^{N \times 128},\\
    f_{\text{conv}} = \textbf{SConvNet}(f_{\text{p}}),\ f_{\text{conv}} \in \mathbb{R}^{N \times 64}.
\end{gathered}
\end{equation}

The hierarchical features are concatenated along the channel dimension and passed to the Noise Predictor $\mathbf{P}$, an MLP network where the output dimension is set to ``1''. For each voxel, the output is a prediction value ranging from 0 to 1, which indicates the level of confidence. The higher the value, the more likely it is to be a valid voxel, and vice versa.
\begin{equation}
    M_{\text{radar}} = \mathbf{P}(h_k \oplus f_{\text{conv}}).
\end{equation}
 $M_{\text{radar}}$ denotes the predicted mask representing positions of valid radar voxel clouds. The mask size is the same as the voxel cloud size.

 The feature extracted from the final layer of the SConvNet module is projected into the BEV representation with feature dimension reconstruction. The BEV representation $f_{\text{BEV}}$ is subsequently processed by the 2D backbone network and detection head. Finally, the detection head outputs category prediction $c_{\text{cls}}$ and box prediction $y_{\text{reg}}$. We use the binary cross-entropy loss 
$\text{BCE}(\cdot)$ for classification and the mean squared error $\text{MSE}(\cdot)$ for regression to supervise the network training:
\begin{equation}
\begin{gathered}
L_{c l s}=\text{BCE} \left(c_{\mathrm{cls}}, c_{\mathrm{gt}}\right), \\
L_{r e g}=\text{MSE} \left(y_{\mathrm{reg}}, y_{\mathrm{gt}} \right),
\end{gathered}
\end{equation}
where $y_{\text{gt}}$ represents the ground truth bounding box, $c_{\text{gt}}$ is the ground truth category of the target area.

\subsection{Cross-modal Supervision Mechanism} \label{sub:cross_modal_supervision_mechanism}
Inspired by prior cross-modal learning methodologies~\cite{yan20222dpass, zhang2023delivering, bang2024radardistill}, our framework introduces a novel supervision mechanism that leverages LiDAR's superior capability in characterizing object features to enhance radar-based target perception. Specifically, LiDAR data is employed during training to supervise the denoising process of 4D radar data while eliminating dependence on LiDAR input during inference. This design enhances radar perception robustness while maintaining practicality without LiDAR.

In the training stage, the data of LiDAR and radar are jointly provided as inputs. The LiDAR data is processed through the Mask Generator, which accomplishes cross-modal matching through a KDTree structure. Specifically, given the radar data $P=\left\{p_i \in \mathbb{R}^3\right\}_{i=1}^{\text{N}}$ and the LiDAR data $Q=\left\{q_j \in \mathbb{R}^3\right\}_{j=1}^{\text{M}}$, the KDTree is constructed on $Q$ with axis-aligned spatial partitioning. At each tree level, the splitting axis alternates cyclically (e.g., $x \rightarrow y \rightarrow z$) to balance spatial discrimination. For each query point $p_i$, the nearest-neighbor search with a fixed distance is performed on $Q$ to identify valid correspondences:
\begin{equation}
N\left(p_i\right)=\left\{q_j \in Q \mid\left\|p_i-q_j\right\|_2<\tau\right\},
\end{equation}
 where $\tau$ is a distance threshold. The binary mask $ {M}_{\text{gt}} $ is then generated as:
\begin{equation}
M_{\text{gt}}(i)= \begin{cases}1 & \text { if }\left|N\left(p_i\right)\right| \geq 1 \\ 0 & \text { otherwise }\end{cases}.
\end{equation}
 This ensures that only points in $P$ with at least one correspondence in $Q$ within $\tau$ are retained, effectively aligning cross-modal features through efficient spatial indexing. The low computation complexity of KDTree enables fast processing for large-scale LiDAR data.

Meanwhile, the radar voxels are passed through the HMSD-Net, where it generates a predicted mask $ {M}_{\text{radar}} $. The two masks are then compared through the loss function: \label{text:kdtree}
\begin{equation}\label{eq:hm_loss}
\textit{L}_{\text{msk}}=\text{SLL}\left({M}_{\text{radar}}, {M}_{\text{gt}}\right),
\end{equation}
where $\text{SSL}$ denotes the Smooth $\text{L}_1$ Loss~\cite{liu2021adaptive}.

During the inference stage, due to the effective denoising capability of the trained denoising network, LiDAR data is no longer required. By removing the need for LiDAR data during inference, this cross-modal framework maintains the efficiency and practicality of radar-based perception while benefiting from the superior precision of LiDAR supervision during training.

\subsection{Total Loss Function}

For the network training of the 3D object detection model, we propose a loss function defined as Equation \eqref{eq:total_loss}:
\begin{equation}\label{eq:total_loss}
    L_{\text {total }}= \alpha \cdot L_{\text {cls}}+\beta \cdot L_{\text {reg }}+\gamma \cdot L_{\text {msk}}.
\end{equation}

\(L_{\text {total }}\) consists of three parts: the classification loss (\(L_{\text {cls}}\)), the regression loss (\(L_{\text {reg}}\)), and the effective voxel loss computed based on the predicted mask and supervised mask (\(L_{\text {msk}}\)). The $L_{\text {cls}}$ utilizes focal loss. The $L_{\text {reg}}$ and $L_{\text {msk}}$ employ the Smooth L1 function. The coefficients, \(\alpha\), \(\beta\), and \(\gamma\) are set as hyperparameters. We utilize the gradient descent algorithm to optimize the weights for minimizing the loss.

\begin{table*}[]
\centering
\caption{QUANTITATIVE COMPARISONS ON THE DUAL-RADAR TEST SET.}
\label{tab:my-table1}
\resizebox{\textwidth}{!}{%
\renewcommand{\arraystretch}{1.2}
\begin{tabular}{ccccccccccccccccccc}
\hline
             & \multicolumn{2}{c}{Car}   & \multicolumn{2}{c}{Cyclist} & \multicolumn{2}{c}{Pedestrian} &  \multicolumn{2}{c}{Mean}  &       
             \\ \cmidrule(r){2-3} \cmidrule(r){4-5} \cmidrule(r){6-7} \cmidrule(r){8-9}
Baselines             & 3D AP (\%)     & BEV AP (\%)    & 3D AP (\%)     & BEV AP (\%)    & 3D AP (\%)          & BEV AP (\%)   & 3D AP (\%)             & BEV AP (\%)       \\ \hline
\multicolumn{6}{l}{\textcolor{lightgray}{Baseline Models for Detection}} \\
PointPillars~\cite{lang2019pointpillars} & 31.78       & 38.30       & 0.14         & 0.29         & 0.00           & 0.00          & 10.64 & 12.87 \\
VoxelR-CNN~\cite{deng2021voxel}   & 35.15       & 39.32       & 0.09         & 0.17         & 0.02           & 0.02          & 11.76 & 13.17 \\
RDIoU~\cite{sheng2022rethinking}        & 32.02       & 36.18       & 0.41         & 0.72         & 0.00           & 0.01          & 10.81 & 12.30 \\
CasA-T~\cite{wu2022casa}       & 8.27        & 15.05       & 0.07         & 0.11         & 0.00           & 0.00          & 2.78  & 5.05  \\
CasA-V~\cite{wu2022casa}       & 14.81       & 16.87       & 0.04         & 0.08         & 0.01           & 0.02          & 4.96  & 5.66  \\ \hline
\multicolumn{6}{l}{\textcolor{lightgray}{Denoising-Enhanced Models}} \\
CasA-V + ROR~\cite{shamsfakhr2024multi} & 41.18       & 45.27       & 3.03         & 3.52         & 0.76           & 0.76          & 14.99 & 16.51 \\
CasA-V + SOR~\cite{xu2023mmlock} & {\ul 46.81} & {\ul 51.18} & 2.15         & 2.47         & 0.91           & 1.14          & 16.62 & 18.26 \\ \hline
\multicolumn{6}{l}{\textcolor{lightgray}{CORENet-Integrated Models}} \\
Voxel R-CNN + CORENet & 46.48          & 51.09          & {\ul 10.02}    & {\ul 10.12}    & {\ul \textbf{1.82}} & \textbf{9.09} & {\ul \textbf{19.44}} & {\textbf {23.43}}    \\
CasA-V + CORENet      & \textbf{49.46} & \textbf{54.38} & \textbf{10.11} & \textbf{10.49} & \textbf{4.55}       & {\ul 4.55}    & \textbf{21.37}       & {\ul 23.14} \\ \hline
  \multicolumn{19}{p{\textwidth}}{\textbf{Bold} is the best, and {\ul{underline}} is the second best. The rightmost two columns report the mean AP scores aggregated across Car, Cyclist, and Pedestrian categories.}
\end{tabular}%
}
\end{table*}

\section{EXPERIMENT}

\subsection{Datasets and Metrics}
We conducted experiments on the Arbe Phoenix radar data and LiDAR data from the Dual-Radar dataset~\cite{zhang2023dual}. After effective selection, the experimental dataset comprises 9966 frames of point cloud data, with 5100, 2400, and 2466 scenes allocated for training, testing, and validation, respectively. The evaluation is based on the official evaluation metric of the KITTI dataset, specifically the average precision (AP) with three predefined difficulty levels: Easy, Moderate, and Hard. To simplify the table, the results under three difficulty levels are averaged to serve as the evaluation metric. The evaluation encompasses metrics from the perspectives of BEV and 3D. In the computation of AP, we employ the Intersection over Union (IoU) thresholds to evaluate the accuracy of detection results. Specifically, an IoU threshold of 50\% is adopted for the Car category, while a threshold of 25\% is applied to the Pedestrian and Cyclist categories.

\subsection{Experimental Setup}
We adopted the CPU of Intel(R) Xeon(R) E5-2650 and built the OpenPCDet framework based on PyTorch 1.8.1. We trained the model on a single RTX 6000 GPU with a 46GB memory capacity using PyTorch. We trained for 80 epochs with a learning rate of 0.003 and presented the best results obtained. The mask generator and match module use a KD tree structure to search for nearest-neighbor matches within 0.5 meters of the input two types of point clouds. Loss hyperparameters \(\alpha\), \(\beta\), and \(\gamma\) are set with values of 1, 2, and 50, respectively. 

\subsection{Baseline Models}
In Table~\ref{tab:my-table1}, baseline models are categorized into two groups for comprehensive comparison. The first group contains advanced detection architectures for point clouds: PointPillars~\cite{lang2019pointpillars} utilizing pillar encoding, Voxel R-CNN~\cite{deng2021voxel}, and RDIoU~\cite{sheng2022rethinking}, along with CasA-T~\cite{wu2022casa} and CasA-V~\cite{wu2022casa}. The second group evaluates denoising methodologies integrated with the CasA-V~\cite{wu2022casa}, including ROR~\cite{shamsfakhr2024multi} and SOR~\cite{xu2023mmlock} filters for outlier removal. Our proposed CORENet is validated through integration with Voxel R-CNN~\cite{deng2021voxel} and CasA-V~\cite{wu2022casa}.

\subsection{ Quantitative Comparison on Dual-Radar}
The quantitative results on the Dual-Radar test set are presented in Table~\ref{tab:my-table1}. Compared with state-of-the-art models and radar-based denoising methods, the improved models perform better in 3D and BEV views. When integrated with CasA-V, CORENet achieves 49.46\% 3D AP and 54.38\% BEV AP for car detection, representing absolute improvements of \textbf{+34.64\%} and \textbf{+37.51\%} over the baseline, respectively. Such a performance gain stems from CORENet's ability to leverage LiDAR-supervised noise patterns to preserve the structural integrity of vehicles in sparse radar data. Similarly, the Voxel R-CNN variant with CORENet attains 46.48\% 3D AP (\textbf{+11.33\%}) and 51.09\% BEV AP (\textbf{+11.78\%}), demonstrating its generalizability across different detection architectures. The superior mean AP results (23.14\% BEV and 21.37\% 3D for CasA-V with CORENet) confirm that explicit cross-modal supervision enables the discriminative separation of valid features from noise clusters.

ROR~\cite{shamsfakhr2024multi} and SOR~\cite{xu2023mmlock} are commonly employed in radar signal processing to enhance point cloud quality in highly noisy point clouds. The proposed CORENet exhibits considerable improvement over the two denoising baselines. For mean AP, CasA-V with CORENet achieves 21.37\% 3D mean AP and 23.14\% BEV mean AP, significantly surpassing ROR (\textbf{+4.75\%} and \textbf{+4.88\%}) and SOR (\textbf{+6.38\%} and \textbf{+6.63\%}).

The quantitative results verify that CORENet enhances detection robustness more effectively than conventional denoising strategies, particularly in preserving geometric fidelity for small objects (cyclists and pedestrians) while maintaining superior car detection accuracy.
\begin{figure*}
  \centering
  \includegraphics[width=0.9\linewidth]{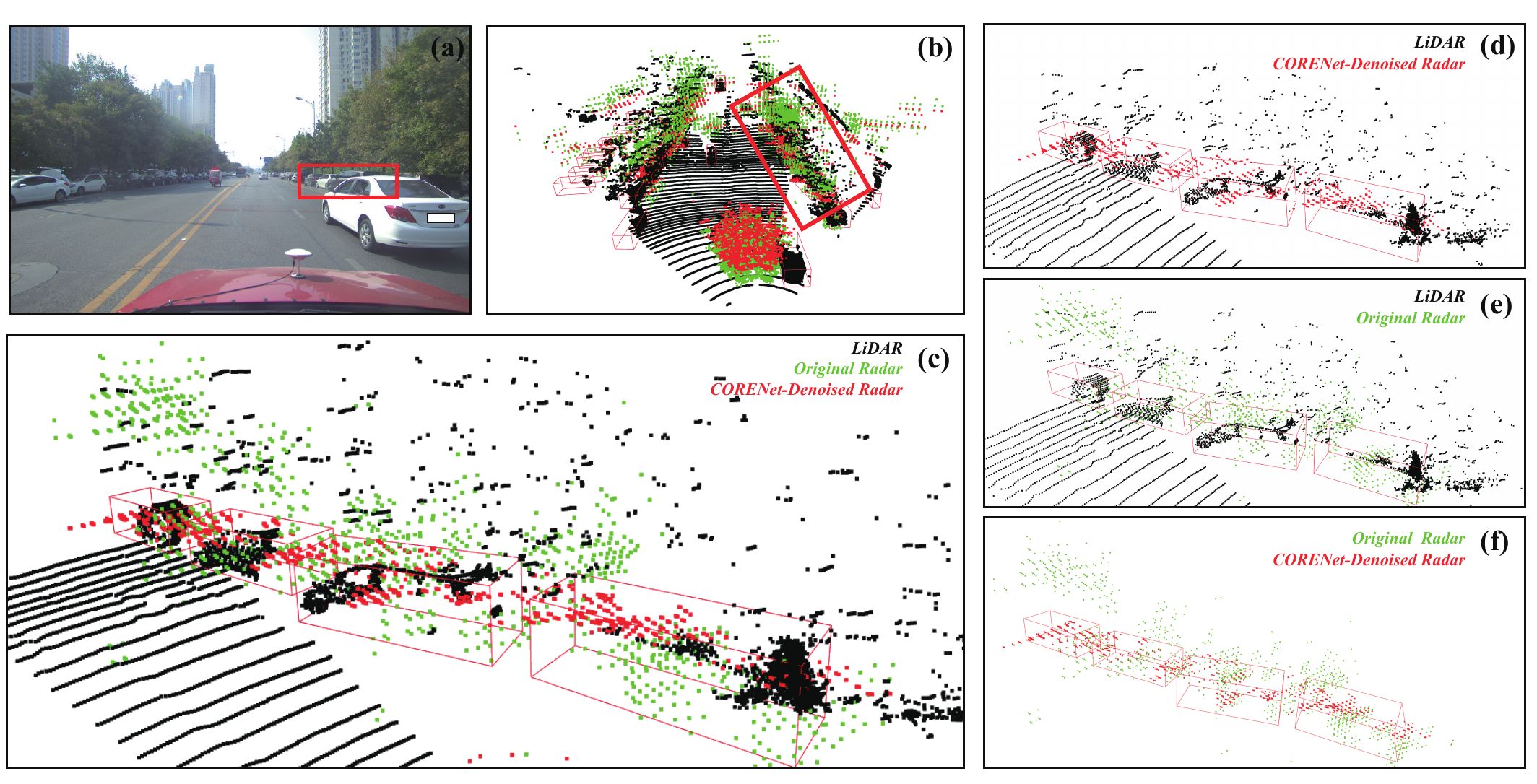}
  \caption{Qualitative results on Dual-Radar dataset. The green and black point clouds correspond to
radar and LiDAR measurements, respectively. The red points are obtained through denoising with CORENet, with red 3D boxes denoting ground truth annotations. A region of interest is highlighted with a red rectangular bounding box. (a) provides a real-world scenario with vehicles on the road. (b) provides an overall visualization of the LiDAR, original 4D radar, and CORENet-denoised radar point clouds. (c-f) offer a zoomed-in view of the area highlighted in the red box from (b).}
  \label{fig:vis_exp}
\end{figure*}

\begin{table}[ht]
\centering
\caption{Ablation Experiments of HMSD-Net}
\label{tab:my-table-lfa}
\resizebox{\columnwidth}{!}{
\begin{tabular}{ccllll}
\\ \hline
\multirow{2}{*}{HPNet} &
  \multirow{2}{*}{SConvNet} &
  \multicolumn{2}{c}{3D AP (\%)} &
  \multicolumn{2}{c}{BEV AP (\%)} \\
  \cmidrule(r){3-4} \cmidrule(r){5-6}
 &
   &
  \multicolumn{1}{c}{Car} &
  \multicolumn{1}{c}{Mean} &
  \multicolumn{1}{c}{Car} &
  \multicolumn{1}{c}{Mean} \\ \hline
$\checkmark$ &           & 67.07 & 26.64 & 70.51 & 28.08 \\
          & $\checkmark$ & 65.94 & 24.92 & 69.43 & 27.64 \\
$\checkmark$ & $\checkmark$  & \textbf{67.34}  & \textbf{27.47}  & \textbf{71.23}   & \textbf{29.01} \\ \hline     
\end{tabular}%
}
\end{table}

\subsection{Ablation Experiments}
To validate the efficacy of the HMSD-Net and the rationality of hyperparameter sets, we conducted ablation experiments on the Dual Radar dataset.

As shown in Table~\ref{tab:my-table-lfa}, we evaluated the contributions of the HPNet and SConvNet components in the HMSD-Net. Compared with the complete model structure, the model with only HPNet resulted in \textbf{-0.72\%} and \textbf{-0.93\%} decline in 3D and BEV mean AP. Using SConvNet alone yielded lower performance, indicating the necessity of combining both components. The HMSD-Net with HPNet and SConvNet achieves the best performance, demonstrating their effectiveness.

\begin{table}[ht]
\centering
\caption{Ablation Experiments of KDTree Distance}
\label{tab:my-table-kdtree}
\resizebox{\columnwidth}{!}{
\begin{tabular}{ccccc}
\hline 
\multirow{2}{*}{\begin{tabular}[c]{@{}c@{}}KDTree \\ Distance/(m)\end{tabular}} & \multicolumn{2}{c}{3D AP (\%)} & \multicolumn{2}{c}{BEV AP (\%)} \\
\cmidrule(r){2-3} \cmidrule(r){4-5}
    & Car        & Mean            & Car       & Mean            \\ \hline
0.3 & 66.94          & 26.86          & 70.26          & 28.00          \\
0.5 & \textbf{67.34} & \textbf{27.47} & \textbf{71.23} & \textbf{29.01} \\
0.7 & 66.71          & 25.81          & 69.70          & 27.11  \\ \hline         
\end{tabular}%
}
\end{table}
\begin{table}[ht]
\centering
\caption{Ablation Experiments of MLP layer in Noise Predictor}
\label{tab:my-table-mlp}
\resizebox{\columnwidth}{!}{
\begin{tabular}{ccccc}
\\ \hline 
\multirow{2}{*}{\begin{tabular}[c]{@{}c@{}}MLP layer in \\ Noise Predictor\end{tabular}} & \multicolumn{2}{c}{3D AP (\%)} & \multicolumn{2}{c}{BEV AP (\%)} \\
\cmidrule(r){2-3} \cmidrule(r){4-5}
  & Car        & Mean            & Car       & Mean            \\ \hline
1 & \textbf{67.34} & \textbf{27.47} & \textbf{71.23} & \textbf{29.01} \\
2 & 66.23          & 26.10          & 69.61          & 27.26          \\
3 & 67.07          & 26.37          & 68.74          & 26.96           \\ \hline 
\end{tabular}%
}
\end{table}

Based on the description in Sec.\ref{text:kdtree}, KDTree structure searches for neighbor voxels in a fixed range of distance. Table~\ref{tab:my-table-kdtree} delineates the models trained with different KDTree distances. A distance of 0.5 yielded the best results, while reducing to 0.3 or increasing to 0.7 caused slight drops. Table~\ref{tab:my-table-mlp} shows the impact of varying MLP layer numbers in the Noise Predictor. The single-layer MLP achieved optimal performance with 67.34\% Car AP and 27.47\% mean AP in 3D detection, along with 71.23\% Car AP and 29.01\% mean AP in BEV detection. This indicates that a single nonlinear transformation suffices to capture noise patterns effectively. However, increasing the depth of layers reduced performance to 66.23\% Car AP (3D) and 69.61\% Car AP (BEV), accompanied by mean AP declines of 1.37\% and 1.75\%, respectively. Notably, a three-layer configuration yielded less decrease in the car category, reaching 67.07\% Car AP and 26.37\% mean AP in 3D, with further degrading BEV performance to 68.74\% Car AP and 26.96\% mean AP (3D), while the results remain inferior compared to the single-layer baseline. The performance degradation with increased depth implies that deeper MLP layers induce overfitting through redundant parameterization, thereby compromising generalization capability. 

\subsection{Visualization and Analysis}

\begin{figure*}[h]
  \centering
  \includegraphics[width=0.9\linewidth]{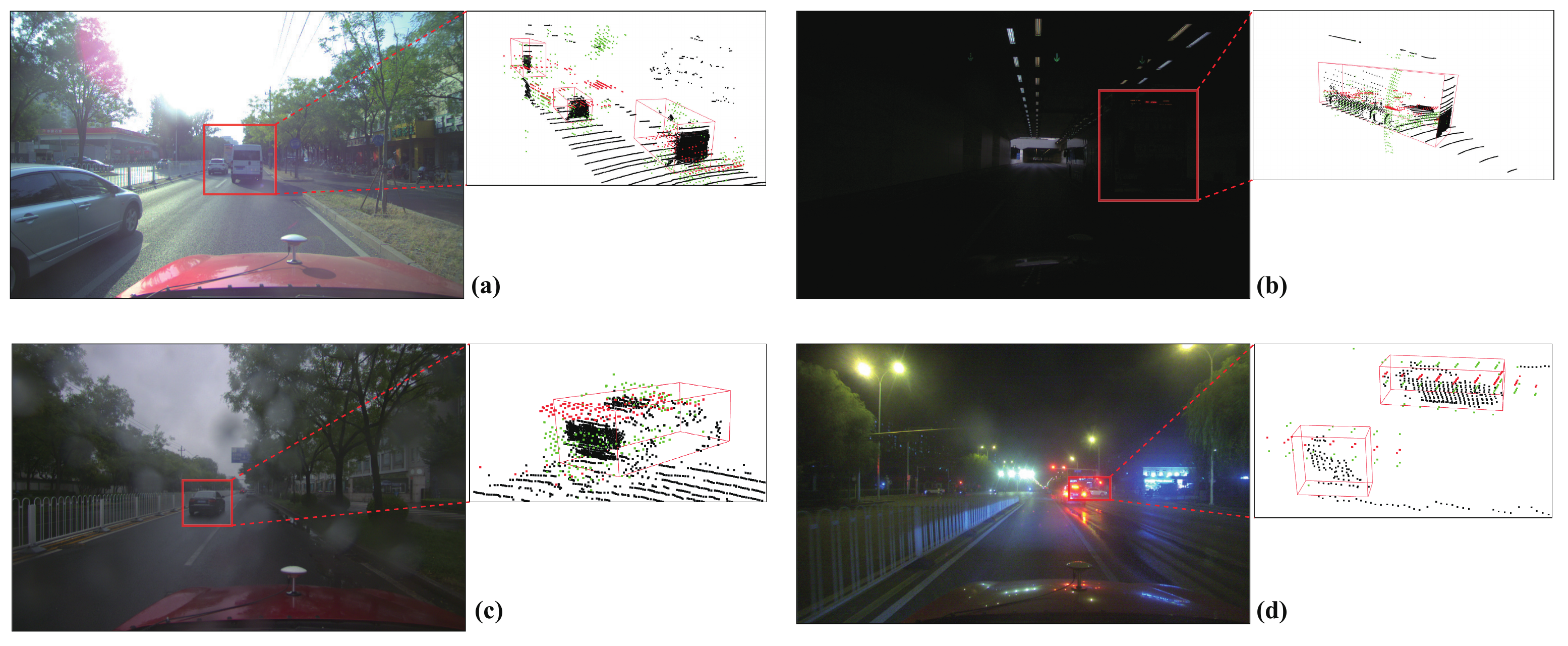}
  \caption{Denoising performance of CORENet in multiple scenes. The green and black point clouds correspond to
radar and LiDAR measurements, respectively. The red points are obtained through denoising with CORENet, with red 3D boxes denoting ground truth annotations. Regions of interest are highlighted with red rectangular boxes and enlarged. (a) daytime; (b) dark tunnel; (c) rainy day; (d) nighttime.}
  \label{fig:vis_exp_2}
\end{figure*}

As illustrated in Fig.~\ref{fig:vis_exp}, we conducted comparisons between scenes before and after denoising, with (d), (e), and (f) presenting pairwise comparisons between different modalities of point clouds. (d) contrasts LiDAR (black) and raw radar (green) point clouds, revealing significant interference induced by radar data noise. High-level noise could greatly impact the detection of objects like vehicles. (e) compares LiDAR (black) and denoised radar data after CORENet (red), indicating that the addition of CORENet provides a point cloud representation closer to the LiDAR ground truth compared to (d), substantially reducing noise in the point cloud. (f) further compares raw radar with CORENet-denoised radar, demonstrating that CORENet effectively reduces radar noise. The denoised point clouds are densely clustered around the vehicle targets and enhance object detection accuracy.

Fig.~\ref{fig:vis_exp_2} demonstrates the multi-scenario denoising performance of CORENet across four distinct environmental conditions: (a) daytime, (b) dark tunnel navigation, (c) rainy day, and (d) nighttime. Each subfigure reveals consistent preservation of structural integrity in detected objects. These qualitative results collectively validate CORENet's cross-environment robustness in preserving spatial information while suppressing noise artifacts, confirming its operational viability for real-world perception systems.

\section{CONCLUSION}
The paper introduced the CORENet as a novel denoising method for 4D radar, which learns to extract valuable point cloud information. Our denoising architecture is simple and effective, allowing seamless integration into mainstream voxel-based detection networks. Additionally, we proposed a cross-modal supervision mechanism that exclusively utilizes LiDAR data during the training phase to enhance performance. Our approach aids the training process, enabling the network to effectively reduce noise in radar point clouds. However, there exist some limitations in the current work. High-precision LiDAR point clouds remain indispensable during the training phase. In future work, we plan to enhance 4D radar detection performance for smaller objects, such as pedestrians and cyclists.

\bibliographystyle{elsarticle-num} 
\bibliography{main.bib}

\begin{thebibliography}{10}
\expandafter\ifx\csname url\endcsname\relax
  \def\url#1{\texttt{#1}}\fi
\expandafter\ifx\csname urlprefix\endcsname\relax\def\urlprefix{URL }\fi
\expandafter\ifx\csname href\endcsname\relax
  \def\href#1#2{#2} \def\path#1{#1}\fi

\bibitem{cui2021deep}
Y.~Cui, R.~Chen, W.~Chu, L.~Chen, D.~Tian, Y.~Li, D.~Cao, Deep learning for
  image and point cloud fusion in autonomous driving: A review, IEEE
  Transactions on Intelligent Transportation Systems 23~(2) (2021) 722--739.

\bibitem{yang2020radarnet}
B.~Yang, R.~Guo, M.~Liang, S.~Casas, R.~Urtasun, Radarnet: Exploiting radar for
  robust perception of dynamic objects, in: Computer Vision--ECCV 2020: 16th
  European Conference, Glasgow, UK, August 23--28, 2020, Proceedings, Part
  XVIII 16, Springer, 2020, pp. 496--512.

\bibitem{Abdu2021ApplicationOD}
F.~J. Abdu, Y.~Zhang, M.~Fu, Y.~Li, Z.~Deng, Application of deep learning on
  millimeter-wave radar signals: A review, Sensors 21~(6) (2021) 1951.

\bibitem{zhang2023dual}
X.~Zhang, L.~Wang, J.~Chen, C.~Fang, L.~Yang, Z.~Song, G.~Yang, Y.~Wang,
  X.~Zhang, J.~Li, Dual radar: A multi-modal dataset with dual 4d radar for
  autononous driving, arXiv preprint arXiv:2310.07602.

\bibitem{amenta2004defining}
N.~Amenta, Y.~J. Kil, Defining point-set surfaces, ACM Transactions on Graphics
  (TOG) 23~(3) (2004) 264--270.

\bibitem{fleishman2005robust}
S.~Fleishman, D.~Cohen-Or, C.~T. Silva, Robust moving least-squares fitting
  with sharp features, ACM transactions on graphics (TOG) 24~(3) (2005)
  544--552.

\bibitem{chen2019multi}
H.~Chen, M.~Wei, Y.~Sun, X.~Xie, J.~Wang, Multi-patch collaborative point cloud
  denoising via low-rank recovery with graph constraint, IEEE transactions on
  visualization and computer graphics 26~(11) (2019) 3255--3270.

\bibitem{yu2022point}
X.~Yu, L.~Tang, Y.~Rao, T.~Huang, J.~Zhou, J.~Lu, Point-bert: Pre-training 3d
  point cloud transformers with masked point modeling, in: Proceedings of the
  IEEE/CVF conference on computer vision and pattern recognition, 2022, pp.
  19313--19322.

\bibitem{du2022dncnet}
M.~Du, P.~Zhong, X.~Cai, D.~Bi, Dncnet: Deep radar signal denoising and
  recognition, IEEE Transactions on Aerospace and Electronic Systems 58~(4)
  (2022) 3549--3562.

\bibitem{shamsfakhr2024multi}
F.~Shamsfakhr, D.~Macii, L.~Palopoli, M.~Corr{\`a}, A.~Ferrari, D.~Fontanelli,
  A multi-target detection and position tracking algorithm based on mmwave-fmcw
  radar data, Measurement 234 (2024) 114797.

\bibitem{xu2023mmlock}
J.~Xu, Z.~Bi, A.~Singha, T.~Li, Y.~Chen, Y.~Zhang, mmlock: User leaving
  detection against data theft via high-quality mmwave radar imaging, in: 2023
  32nd International Conference on Computer Communications and Networks
  (ICCCN), IEEE, 2023, pp. 1--10.

\bibitem{sun20214d}
S.~Sun, Y.~D. Zhang, 4d automotive radar sensing for autonomous vehicles: A
  sparsity-oriented approach, IEEE Journal of Selected Topics in Signal
  Processing 15~(4) (2021) 879--891.

\bibitem{liu2024planar}
H.~Liu, J.~Li, Z.-C. Hao, Y.~Hu, G.~Xu, W.~Hong, A planar millimeter-wave
  diffuse-reflection suppression 4d imaging radar using l-shaped switchable
  linearly phased array, IEEE Transactions on Radar Systems.

\bibitem{yan20222dpass}
X.~Yan, J.~Gao, C.~Zheng, C.~Zheng, R.~Zhang, S.~Cui, Z.~Li, 2dpass: 2d priors
  assisted semantic segmentation on lidar point clouds, in: European Conference
  on Computer Vision, Springer, 2022, pp. 677--695.

\bibitem{zhang2023delivering}
J.~Zhang, R.~Liu, H.~Shi, K.~Yang, S.~Rei{\ss}, K.~Peng, H.~Fu, K.~Wang,
  R.~Stiefelhagen, Delivering arbitrary-modal semantic segmentation, in:
  Proceedings of the IEEE/CVF Conference on Computer Vision and Pattern
  Recognition, 2023, pp. 1136--1147.

\bibitem{bang2024radardistill}
G.~Bang, K.~Choi, J.~Kim, D.~Kum, J.~W. Choi, Radardistill: Boosting
  radar-based object detection performance via knowledge distillation from
  lidar features, in: Proceedings of the IEEE/CVF Conference on Computer Vision
  and Pattern Recognition, 2024, pp. 15491--15500.

\bibitem{palmer2024lerojd}
P.~Palmer, M.~Kr{\"u}ger, S.~Sch{\"u}tte, R.~Altendorfer, G.~Adam, T.~Bertram,
  Lerojd: Lidar extended radar-only object detection, in: European Conference
  on Computer Vision, Springer, 2024, pp. 379--396.

\bibitem{svenningsson2021radar}
P.~Svenningsson, F.~Fioranelli, A.~Yarovoy, Radar-pointgnn: Graph based object
  recognition for unstructured radar point-cloud data, in: 2021 IEEE Radar
  Conference (RadarConf21), IEEE, 2021, pp. 1--6.

\bibitem{shi2020point}
W.~Shi, R.~Rajkumar, Point-gnn: Graph neural network for 3d object detection in
  a point cloud, in: Proceedings of the IEEE/CVF conference on computer vision
  and pattern recognition, 2020, pp. 1711--1719.

\bibitem{ulrich2022improved}
M.~Ulrich, S.~Braun, D.~K{\"o}hler, D.~Niederl{\"o}hner, F.~Faion,
  C.~Gl{\"a}ser, H.~Blume, Improved orientation estimation and detection with
  hybrid object detection networks for automotive radar, in: 2022 IEEE 25th
  International Conference on Intelligent Transportation Systems (ITSC), IEEE,
  2022, pp. 111--117.

\bibitem{kim2023craft}
Y.~Kim, S.~Kim, J.~W. Choi, D.~Kum, Craft: Camera-radar 3d object detection
  with spatio-contextual fusion transformer, in: Proceedings of the AAAI
  Conference on Artificial Intelligence, Vol.~37, 2023, pp. 1160--1168.

\bibitem{kim2023crn}
Y.~Kim, J.~Shin, S.~Kim, I.-J. Lee, J.~W. Choi, D.~Kum, Crn: Camera radar net
  for accurate, robust, efficient 3d perception, in: Proceedings of the
  IEEE/CVF International Conference on Computer Vision, 2023, pp. 17615--17626.

\bibitem{nabati2021centerfusion}
R.~Nabati, H.~Qi, Centerfusion: Center-based radar and camera fusion for 3d
  object detection, in: Proceedings of the IEEE/CVF Winter Conference on
  Applications of Computer Vision, 2021, pp. 1527--1536.

\bibitem{wang2023bi}
Y.~Wang, J.~Deng, Y.~Li, J.~Hu, C.~Liu, Y.~Zhang, J.~Ji, W.~Ouyang, Y.~Zhang,
  Bi-lrfusion: Bi-directional lidar-radar fusion for 3d dynamic object
  detection, in: Proceedings of the IEEE/CVF Conference on Computer Vision and
  Pattern Recognition, 2023, pp. 13394--13403.

\bibitem{zhou2023bridging}
T.~Zhou, J.~Chen, Y.~Shi, K.~Jiang, M.~Yang, D.~Yang, Bridging the view
  disparity between radar and camera features for multi-modal fusion 3d object
  detection, IEEE Transactions on Intelligent Vehicles 8~(2) (2023) 1523--1535.

\bibitem{kim2024rcm}
J.~Kim, M.~Seong, G.~Bang, D.~Kum, J.~W. Choi, Rcm-fusion: Radar-camera
  multi-level fusion for 3d object detection, in: 2024 IEEE International
  Conference on Robotics and Automation (ICRA), IEEE, 2024, pp. 18236--18242.

\bibitem{oztireli2009feature}
A.~C. {\"O}ztireli, G.~Guennebaud, M.~Gross, Feature preserving point set
  surfaces based on non-linear kernel regression, in: Computer graphics forum,
  Vol.~28, Wiley Online Library, 2009, pp. 493--501.

\bibitem{alexa2003computing}
M.~Alexa, J.~Behr, D.~Cohen-Or, S.~Fleishman, D.~Levin, C.~T. Silva, Computing
  and rendering point set surfaces, IEEE Transactions on visualization and
  computer graphics 9~(1) (2003) 3--15.

\bibitem{wang2019dynamic}
Y.~Wang, Y.~Sun, Z.~Liu, S.~E. Sarma, M.~M. Bronstein, J.~M. Solomon, Dynamic
  graph cnn for learning on point clouds, ACM Transactions on Graphics (tog)
  38~(5) (2019) 1--12.

\bibitem{qi2017pointnet}
C.~R. Qi, H.~Su, K.~Mo, L.~J. Guibas, Pointnet: Deep learning on point sets for
  3d classification and segmentation, in: Proceedings of the IEEE conference on
  computer vision and pattern recognition, 2017, pp. 652--660.

\bibitem{qi2017pointnet++}
C.~R. Qi, L.~Yi, H.~Su, L.~J. Guibas, Pointnet++: Deep hierarchical feature
  learning on point sets in a metric space, Advances in neural information
  processing systems 30.

\bibitem{liu2021deep}
S.-L. Liu, H.-X. Guo, H.~Pan, P.-S. Wang, X.~Tong, Y.~Liu, Deep implicit moving
  least-squares functions for 3d reconstruction, in: Proceedings of the
  IEEE/CVF Conference on Computer Vision and Pattern Recognition, 2021, pp.
  1788--1797.

\bibitem{de2023iterativepfn}
D.~de~Silva~Edirimuni, X.~Lu, Z.~Shao, G.~Li, A.~Robles-Kelly, Y.~He,
  Iterativepfn: True iterative point cloud filtering, in: Proceedings of the
  IEEE/CVF Conference on Computer Vision and Pattern Recognition, 2023, pp.
  13530--13539.

\bibitem{deng2021voxel}
J.~Deng, S.~Shi, P.~Li, W.~Zhou, Y.~Zhang, H.~Li, Voxel r-cnn: Towards high
  performance voxel-based 3d object detection, in: Proceedings of the AAAI
  conference on artificial intelligence, Vol.~35, 2021, pp. 1201--1209.

\bibitem{zhou2021adaptive}
H.~Zhou, Y.~Feng, M.~Fang, M.~Wei, J.~Qin, T.~Lu, Adaptive graph convolution
  for point cloud analysis, in: Proceedings of the IEEE/CVF international
  conference on computer vision, 2021, pp. 4965--4974.

\bibitem{graham20183d}
B.~Graham, M.~Engelcke, L.~Van Der~Maaten, 3d semantic segmentation with
  submanifold sparse convolutional networks, in: Proceedings of the IEEE
  conference on computer vision and pattern recognition, 2018, pp. 9224--9232.

\bibitem{liu2021adaptive}
C.~Liu, S.~Yu, M.~Yu, B.~Wei, B.~Li, G.~Li, W.~Huang, Adaptive smooth l1 loss:
  A better way to regress scene texts with extreme aspect ratios, in: 2021 IEEE
  Symposium on Computers and Communications (ISCC), IEEE, 2021, pp. 1--7.

\bibitem{lang2019pointpillars}
A.~H. Lang, S.~Vora, H.~Caesar, L.~Zhou, J.~Yang, O.~Beijbom, Pointpillars:
  Fast encoders for object detection from point clouds, in: Proceedings of the
  IEEE/CVF conference on computer vision and pattern recognition, 2019, pp.
  12697--12705.

\bibitem{sheng2022rethinking}
H.~Sheng, S.~Cai, N.~Zhao, B.~Deng, J.~Huang, X.-S. Hua, M.-J. Zhao, G.~H. Lee,
  Rethinking iou-based optimization for single-stage 3d object detection, in:
  European Conference on Computer Vision, Springer, 2022, pp. 544--561.

\bibitem{wu2022casa}
H.~Wu, J.~Deng, C.~Wen, X.~Li, C.~Wang, J.~Li, Casa: A cascade attention
  network for 3-d object detection from lidar point clouds, IEEE Transactions
  on Geoscience and Remote Sensing 60 (2022) 1--11.

\end{thebibliography}

\end{document}